\documentclass[conference]{IEEEtran}
\ifCLASSINFOpdf
  \usepackage[pdftex]{graphicx}
\else
\fi
\usepackage[cmex10]{amsmath}
\usepackage{amsfonts}

\newcommand{\Acal}{\mathcal{A}}
\newcommand{\Bcal}{\mathcal{B}}
\newcommand{\Ccal}{\mathcal{C}}

\newcommand{\Jcal}{\mathcal{J}}

\newcommand{\Pcal}{\mathcal{P}}
\newcommand{\Qcal}{\mathcal{Q}}

\newcommand{\Scal}{\mathcal{S}}

\newcommand{\Ucal}{\mathcal{U}}
\newcommand{\Vcal}{\mathcal{V}}
\newcommand{\Xcal}{\mathcal{X}}
\newcommand{\Ycal}{\mathcal{Y}}

\newcommand{\Rbb}{\mathbb{R}}

\newcommand{\bs}{\boldsymbol}

\newtheorem{definition}{Definition}
\newtheorem{theorem}{Theorem}

\usepackage[tight,footnotesize]{subfigure}

\usepackage{float} 
\begin{document}
%
\title{Image classification using local tensor singular value decompositions}

\author{\IEEEauthorblockN{Elizabeth Newman}
\IEEEauthorblockA{Department of Mathematics\\
Tufts University\\
Medford, Massachusetts 02155\\
Email: e.newman@tufts.edu}
\and
\IEEEauthorblockN{Misha Kilmer}
\IEEEauthorblockA{Department of Mathematics\\
Tufts University\\
Medford, Massachusetts 02155\\
Email: misha.kilmer@tufts.edu}
\and
\IEEEauthorblockN{Lior Horesh}
\IEEEauthorblockA{IBM TJ Watson Research Center\\
1101 Kitchawan Road\\
Yorktown Heights, NY\\
Email: lhoresh@us.ibm.com}
}


%


\maketitle

\begin{abstract}

From linear classifiers to neural networks, image classification has been a widely explored topic in mathematics, and many algorithms have proven to be effective classifiers.  However, the most accurate classifiers typically have significantly high storage costs, or require complicated 
procedures 
that may be computationally expensive.  We present a novel (nonlinear) classification approach using truncation of local tensor singular value decompositions (tSVD)
that robustly offers accurate results, while maintaining manageable storage costs. 
Our approach takes advantage of the optimality of the representation under the tensor algebra described to determine to which class an image belongs.  We extend our approach to a method that can determine specific pairwise match scores, which could be useful in, for example, object recognition problems where pose/position are different.  We 
demonstrate 
the promise of our new techniques on the MNIST data set.
\end{abstract}


%
\IEEEpeerreviewmaketitle

\section{Introduction}

Image classification is a well-explored problem in which an image is identified as belonging to one of a known number of classes.  Researchers seek to extract particular features from which to determine 
patterns comprising an image. 
 Algorithms to determine these essential features include statistical methods such as
centroid-based clustering, connectivity
/graph-
based clustering, distribution-based clustering, and density-based clustering \cite{Rokach, Lloyd, Ester}, as well as
learning algorithms (linear discriminant analysis, support vector machines, neural networks) \cite{Theodoris2006}.  

Our approach differs significantly from techniques in the literature in that it uses local tensor singular value decompositions (tSVD) 
to form the feature 
space 
 of an image.  
 Tensor approaches are gaining increasing popularity for tasks such as image recognition and dictionary learning and reconstruction \cite{Hao2013, Mujamder, Alex, Soltani}.  These are favored over matrix-vector-based approaches as it has been demonstrated that a tensor-based approach enables retention of the original image 
 structural correlations 
 that are lost by image vectorization.  
 Tensor approaches for image classification appear to be in their infancy, although some approaches based on the tensor HOSVD \cite{Lieven} have been explored in the literature \cite{SavasElden}. 
 
Here, we are motivated by the work in \cite{Hao2013} which employs optimal low tubal-rank tensor factorizations through use of the t-product \cite{KilmerMartin} and by the work in \cite{Kilmer2015} describing tensor orthogonal projections.  We present 
a new 
approach for classification based on the tensor SVD from \cite{KilmerMartin}, called the tSVD, which is elegant for its straightforward mathematical interpretation and implementation, and which has the advantage that it can be {\it easily parallelized for great computational advantage}.  
  State-of-the-art matrix decompositions are asymptotically challenged in dealing with the demand to process ever-growing datasets of larger and more complex objects 
 \cite{Boutsidis}, so the importance of this 
 dimension 
 of 
 this study 
 cannot be overstated.      
 Our method is in direct contrast to 
 deep 
 neural network based approaches which require many layers of complexity and for which theoretical interpretation is not readily available \cite{Shalev-Shwartz}.  Our approach is also different from the tensor approach in \cite{SavasElden} because truncating the tSVD has optimality properties that truncating the HOSVD does not enjoy.  We conclude this 
 study 
 with a demonstration on the MNIST \cite{MNIST} dataset.        


\subsection{Notation and Preliminaries}\label{subsec:notation}

\begin{figure}[H]
\centering
\subfigure[Tensor $\Acal$.]{\includegraphics[scale=0.13]{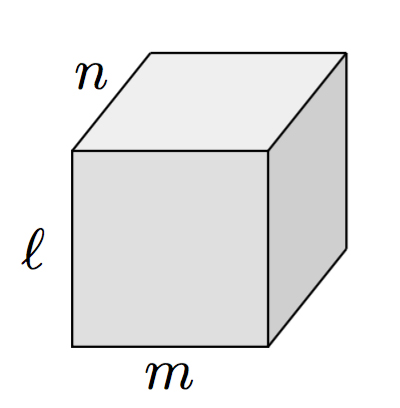}}
\subfigure[Frontal \newline \hspace*{0.35cm}  slices $A^{(k)}$.]{\includegraphics[scale=0.13]{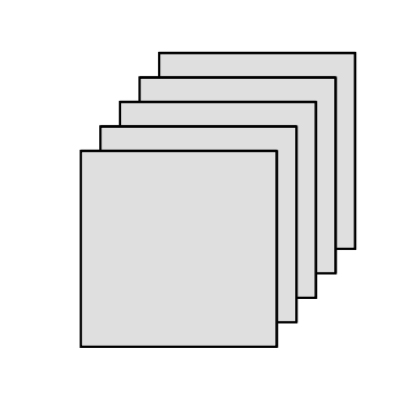}}
\subfigure[Lateral \newline \hspace*{0.35cm} slices $\vec{\Acal}_j$.]{\includegraphics[scale=0.13]{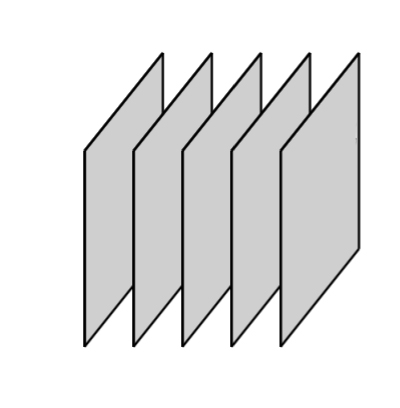}} \subfigure[Tubes $\bs a_{ij}$.]{\includegraphics[scale=0.13]{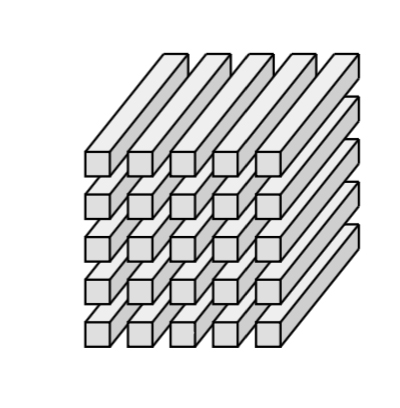}}
\caption{Representations of third-order tensors.}
\label{fig:tensor}
\end{figure}

In this paper, a \emph{tensor} is a third-order tensor, or three-dimensional array of data, denoted by a capital script letter.  As depicted in Figure \ref{fig:tensor}, $\Acal$ is an $\ell \times m\times n$ tensor.  \emph{Frontal slices} $A^{(k)}$ for $k = 1,\dots, n$ are $\ell\times m$ matrices.  \emph{Lateral slices} $\vec{\Acal}_j$ for $j = 1,\dots, m$ are $\ell\times n$ matrices oriented along the third dimension.  \emph{Tubes} $\bs{a}_{ij}$ for $i = 1,\dots, \ell$ and $j = 1,\dots, m$ are $n\times 1$ column vectors oriented along the third dimension \cite{Kilmer2015}. 

To paraphrase the definition by Kilmer et al. \cite{Kilmer2015}, the range of a tensor $\Acal$ is the \emph{t-linear span} of the lateral slices of $\Acal$:
	\begin{equation}\label{defn:range1}
	R(\Acal) = \{\vec{\Acal}_1 *\bs c_ 1 + \dots  + \vec{\Acal}_m * \bs c_m\mid \bs c_i \in \Rbb^{1\times1\times n}\}.
	\end{equation}
Because the lateral slices of $\vec{\Acal}$ form the range, we store our images as lateral slices. Furthermore, $\Acal$ is real-valued because images are real-value.

To multiply a pair of tensors, we need to understand the t-product, which requires the following tensor reshaping 
machinery. 
 Given $\Acal\in \Rbb^{\ell\times m\times n}$, the \texttt{unfold} function reshapes $\Acal$ into an $\ell n\times m$ block-column vector (ie. the first block-column of (\ref{defn:bcirc})), while \texttt{fold} folds it back up again.   
The $\texttt{bcirc}$ function forms an $\ell n\times mn$ block-circulant matrix from the frontal slices of $\Acal$: 
	\begin{equation}\label{defn:bcirc}
	\texttt{bcirc}(\Acal) = \begin{pmatrix} 
					A^{(1)} & A^{(n)} & \dots & A^{(2)}\\
					A^{(2)} & A^{(1)} & \dots & A^{(3)}\\
					\vdots & \vdots & \ddots & \vdots \\
					A^{(n)} & A^{(n-1)} & \dots & A^{(1)}
					\end{pmatrix}.
	\end{equation}

Now the t-product is defined as follows (\cite{KilmerMartin}):
\begin{definition}[t-product]\label{defn:tprod}
Given $\Acal \in \Rbb^{\ell\times p\times n}$ and $\Bcal \in \Rbb^{p\times m \times n}$, 
the \emph{t-product} is the $\ell \times m \times n$ product
	\begin{equation}\label{eqn:tprod}
	\Acal * \Bcal = \texttt{fold}(\texttt{bcirc}(\Acal) \cdot \texttt{unfold}(\Bcal)).
	\end{equation}
\end{definition}

Under the t-product (Definition \ref{defn:tprod}), we need the following from \cite{KilmerMartin}.
	

\begin{definition}\label{defn:transpose}
The \emph{tensor transpose} $\Acal^T\in \Rbb^{p\times \ell \times n}$  takes the transpose of the frontal slices of $\Acal$ and reverses the order of slices $2$ through $n$.
\end{definition}	

\begin{definition}\label{defn:identity}
The \emph{identity tensor} $\Jcal$ is an $m\times m \times n$ tensor 
where $\Jcal^{(1)}$ an $m\times m$ identity matrix and all other frontal slices are zero.
\end{definition}

\begin{definition}\label{defn:orthogonal}
An \emph{orthogonal tensor} $\Qcal$ is an $m\times m \times n$ tensor such that $\Qcal^T * \Qcal = \Qcal * \Qcal^T = \Jcal.$  
\end{definition}
Analogous to the columns of an orthogonal matrix, the lateral slices of $\Qcal$ are orthonormal \cite{Kilmer2015}.

\begin{definition}\label{defn:f-diagonal}
A tensor is \emph{f-diagonal} if each frontal slice is a diagonal matrix.
\end{definition}

%
%

\section{Tensor Singular Value Decomposition}

Let $\Acal$ be an $\ell\times m \times n$ tensor.  As defined in \cite{KilmerMartin}, the \emph{tensor singular value decomposition (tSVD)} of $\Acal$ is the following:
	\begin{align}\label{defn:tsvd}
	\Acal &= \Ucal * \Scal * \Vcal^T,
	\end{align}
where for $p = \min(\ell,m)$, $\Ucal$ is an $\ell\times p \times n$ tensor with orthonormal lateral slices, $\Vcal$ is a $m\times p\times n$ tensor with orthonormal lateral slices, and $\Scal$ is $p\times p \times n$ f-diagonal.

The algorithm for computing the tSVD is given in \cite{KilmerMartin}.  Importantly, as noted in that paper, the bulk of the computations are 
performed 
on matrices, which are {\it independent and can thus be done in parallel}.   Furthermore, 
synonymously to matrix computation strategies, 
randomized variants of the tSVD algorithm have recently been proposed \cite{Jiani} which can be favored when the tensor is particularly large.  

\subsection{Range and Tubal-Rank of Tensors}

As proven in Kilmer et al. \cite{Kilmer2015}, the \emph{range} of $\Acal$ determined via t-linear combinations of the lateral slices of $\Ucal$, for appropriate tensor coefficients $\bs c_i$:
	\begin{equation}\label{defn:range}
	R(\Acal) = \{\vec{\Ucal}_1 * \bs c_1 + \dots + \vec{\Ucal}_p * \bs c_p \mid \bs c_i \in \Rbb^{1\times 1\times n}\}.
	\end{equation}
The lateral slices of $\Ucal$ form an orthonormal basis for the range of $\Acal$.  More details related to the definition and the rest of the linear-algebraic framework can be found in \cite{Kilmer2015}. 
 
 The definition of the range of a tensor leads to the notion of projection.  Given a lateral slice $\vec{\Bcal}\in \Rbb^{m\times 1\times n}$, the \emph{orthogonal projection} into the range of $\Acal$ is defined as $\Ucal*\Ucal^T * \vec{\Bcal}$.


We require the following theorem to understand tubal-rank of tensors:

\begin{theorem}[ \cite{KilmerMartin}]\label{thm:truncation tsvd}
For $k\le \min(\ell,m)$, define
	\begin{align*}
	\Acal_k &= \sum_{i=1}^k \vec{\Ucal}_i * \bs{s}_{ii} * \vec{\Vcal}_i^T.
	\end{align*}
	where $\vec{\Ucal}_i$ and $\vec{\Vcal}_i$ are the $ith$ lateral slices of $\Ucal$ and $\Vcal$, respectively, and $\bs{s}_{ii}$ is the $(i,i)$-tube of $\Scal$.
Then $\Acal_k = \text{arg}\min_{\tilde{\Acal}\in M} || \Acal - \tilde{\Acal}||_F$ where $M = \{\Ccal = \Xcal * \Ycal \mid \Xcal \in \Rbb^{\ell\times k\times n}, \Ycal \in \Rbb^{k\times m\times n}\}$.
\end{theorem}

From Theorem~\ref{thm:truncation tsvd}, we say $\Acal_k$ is a tensor of tubal-rank-$k$. The definition of tubal rank is from \cite{Kilmer2015}.  
It follows from the above that $\Acal_k$ is {\it best tubal-rank-$k$ approximation} to $\Acal$.

\subsection{The Algorithm}

Suppose we have a set of training images and each image in the set belongs to one of $N$ different classes.  First, we form a third-order tensor\footnote{ We note that extensions of the t-product and corresponding decompositions are possible for higher order tensor representations (e.g. for color image training data), as well \cite{HaoHoreshKilmer, MartinLaRue}.} for each class $\Acal_1,\Acal_2,\dots, \Acal_N$ where $\Acal_i$ contains all the training images belonging to class $i$, stored as lateral slices.  We assume all the training images are $\ell \times n$ and that there are $m_i$ images in class $i$; i.e., $\Acal_i$ is an $\ell \times m_i \times n$ tensor.  Note that the $m_i$ need not be the same.   We then form a tubal-rank-$k$ \emph{local tSVD} (Theorem~\ref{thm:truncation tsvd}) for each tensor:
\begin{align}\label{defn:local tsvd}
	\Acal_i &\approx \Ucal_i * \Scal_i * \Vcal_i^T  \text{ for } i = 1,\dots, N,
	\end{align}
where $\Ucal_{i}$ is an $\ell\times k \times n$ tensor.   Here, $k \ll m_i$.  Now, instead of storing all the training images, we need only store an $\ell \times k \times n$ tensor for each class.    The training basis is thus an optimal basis in the sense of Theorem~\ref{thm:truncation tsvd}.    The tensor operator $ \Ucal_i  *\Ucal_i^T$ is an orthogonal projection tensor \cite{Kilmer2015} onto the space which is the t-linear combination of the lateral slices of the $\Ucal_i$ tensor.   Likewise, $ (\mathcal I - \Ucal_i * \Ucal_i^T)$ projects orthogonally to this space.   

Next, suppose a test image belongs to one of the $N$ classes and we want to determine the class to which it belongs.  We re-orient this image as a lateral slice $\vec{\Bcal}$ and use our local tSVD bases 
to compute the norms of the tensor coefficients of the image projected orthogonally to the current training set:
	\begin{equation}\label{defn:proj metric}
	\text{arg}\min_{i=1,\dots,N} ||\vec{\Bcal} - \Ucal_i * \Ucal_i^T * \vec{\Bcal} ||_F, \text{ for } i=1,\ldots,N.
	\end{equation}
If $\vec{\Bcal}$ is a member of the class $i$, 
we expect \eqref{defn:proj metric} to be small.  We determine the class to which $\vec{\Bcal}$ belongs by which projection is the closest to the original image in the Frobenius norm.


\section{Experiments and Results}

To test our local tSVD classifier, we use the public MNIST dataset of handwritten digits as a benchmark~\cite{MNIST}.  The MNIST dataset contains of $60,000$ training images and $10,000$ test images.  Each image is a $28\times 28$ grayscale image consisting of a single hand-written digit (i.e., $0$ through $9$).  We organize the training images by digit resulting in $10$ different classes with the distribution of digits displayed in Figure~\ref{fig:digit distribution}.
	\begin{figure}[H]
	\centering
	\caption{Table of MNIST digit distribution.}
	\vspace{0.15cm}
	
	\begin{tabular}{|l|c|c|c|c|c|}
	\hline
	Digit & 0 & 1 & 2 & 3 & 4\\
	\hline
	\# training & 5923 & 6742&5958 &6131 &5842 \\
	\# test & 980 & 1135&1032 &1010 &982 \\
	\hline
	\end{tabular}
	
	\vspace{0.25cm}
	\centering
	\hspace{0.13cm}\begin{tabular}{|l|c|c|c|c|c|}
	\hline
	Digit & 5 & 6 & 7 & 8 & 9\\
	\hline
	\# training & 5421 & 5918&6265 &5851 &5949 \\
	\# test & 892 & 958&1028 &974 &1009 \\
	\hline
	\end{tabular}
	\label{fig:digit distribution}
	\end{figure}

We store each class of training images as a tensor with the images stored as lateral slices (e.g., the tensor containing images of the digit $0$ is of size $28 \times 5923 \times 28$).  Using \eqref{defn:local tsvd}, we independently form a local tSVD basis for each class $\Ucal_0,\Ucal_1,\dots, \Ucal_9$ where $\Ucal_i$ is the basis for the digit $i$ and of size $28 \times k \times 28$ for some truncation $k$.  For simplicity, we use the same truncation for all bases \footnote{ Note that the tSVD offers flexibility in prescription of the truncation level per basis \cite{Hao2013}.   }. 

\subsection{Numerical Results:  Classification}

Our first objective is to use these local tSVD bases to determine the digit in each test image.  Suppose $\vec{\Pcal}_j$ is the $28 \times 1\times 28$ lateral slice of the $jth$ test image.  We determine how similar $\vec{\Pcal}_j$ is to each digit using the following metric \eqref{defn:proj metric}:
	\begin{equation}\label{defn:projection}
	\text{arg}\min_{i=0,\dots,9} ||\vec{\Pcal}_j - \Ucal_i* \Ucal_i^T * \vec{\Pcal}_j ||_F.
	\end{equation}

To measure the accuracy of our classification, we compute the recognition rate for the entire test data as follows:
	\begin{equation}\label{defn:rec rate}
	r = \dfrac{\text{\# of correctly classified test images}}{\text{\# of test images}}.
	\end{equation}

For various truncation values $k$, we obtain the following recognition rates:

\begin{figure}[H]
\centering
\caption{Classification accuracy for various truncation values.}
\begin{tabular}{|r|c|c|c|c|}
\hline
Truncation	& $k = 3$ & $k = 4$ & $k = 5$ & $k = 10$\\
	\hline
$r$ (\%) &87.99 & 88.51  & 87.14 & 75.31\\
\hline
\end{tabular}

\label{fig:total rec rate}
\end{figure}

\begin{figure}[H]
	\centering
	{\includegraphics[scale=0.15]{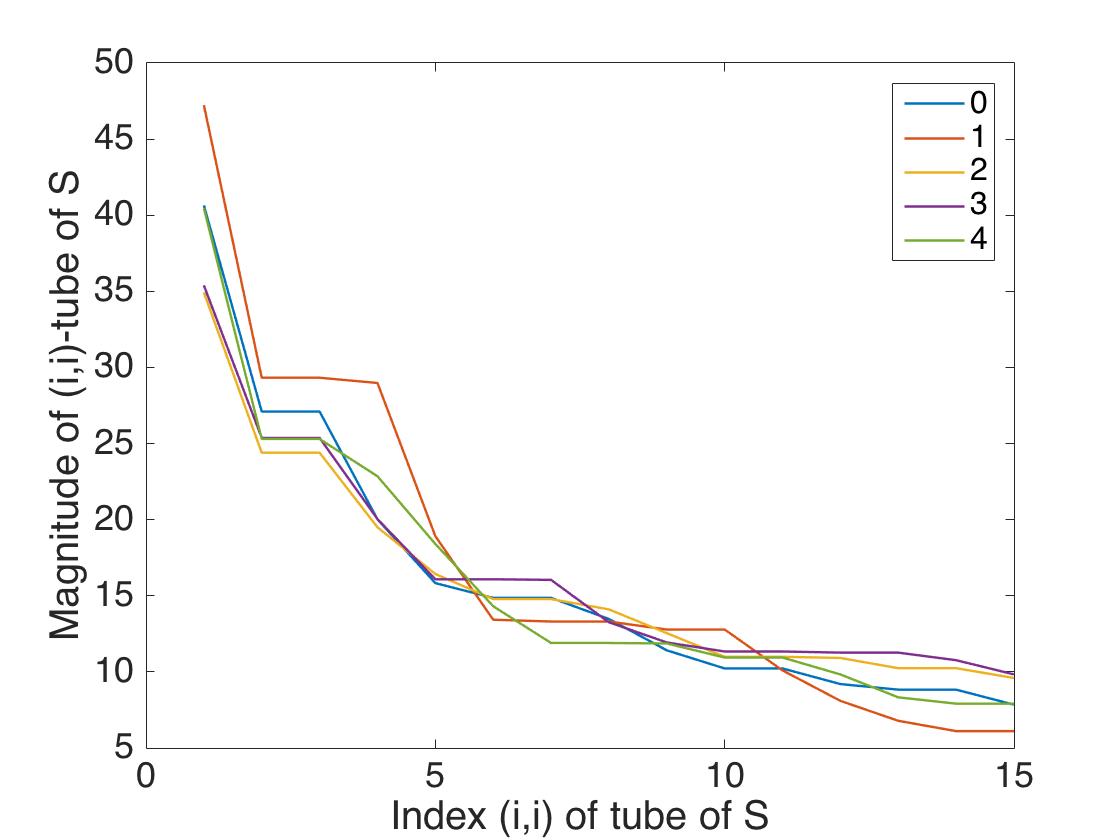}}
	\caption{Magnitude decay of norm of singular value tubes for digits 0-4.}
	\label{fig:svalues}
	\end{figure}

From Figure~\ref{fig:total rec rate}, we notice that smaller truncation values yield greater classification accuracy.  This indicates that the magnitude of the tubes of singular values in $\Scal$ (i.e., $||\bs s_{ii}||_F$) decays rapidly for the early truncation values, as demonstrated in Figure~\ref{fig:svalues}.
\vspace{-0.25cm}


Notice in Figure~\ref{fig:svalues} the magnitude of the tubes of $\Scal$ decays rapidly for the first few indices $i$ and decays more slowly starting at the index $i = 5$.  
This 
implies 
we can optimize our storage costs by truncating at about $k = 5$ without losing significant classification accuracy.  


In addition 
to the overall classification accuracy, we can measure the accuracy of classifying each digit as 
		\begin{equation}\label{defn:rec rate digit}
	r_i = \dfrac{\text{\# of correctly classified test images of digit $i$}}{\text{\# of test images of digit $i$}}.
	\end{equation}
	
We show the per-digit accuracy results for $k=4$ below:
\begin{figure}[H]
\centering
\caption{Classification accuracy per digit for truncation $k = 4$.}
\begin{tabular}{|c||c|c|c|}
\hline
Digit & Most Freq. & 2nd Most& $r_i$ (\%)\\
\hline\hline
0 &0 & 1&91.12\\
\hline
1 &1 &4 & 96.56\\
\hline
2 & 2& 0&83.92\\
\hline
3 & 3&8 &82.77\\
\hline
4 & 4&1 &96.13\\
\hline
5 &5 &8 &79.48\\
\hline
6 &6 &1 &93.32\\
\hline
7 &7 &9 &90.95\\
\hline
8 & 8&5 &82.14\\
\hline
9 &9 & 4&87.02\\
\hline
\end{tabular}
\label{fig:classification}
\end{figure}

In Figure~\ref{fig:classification}, the ``Most Freq.'' column indicates the class to which the images of each digit were most frequently classified.  The ``2nd Most'' column indicates the second class to which the images of each digit were most frequently classified.  

\begin{figure}[H]
\centering
\subfigure[Incorrectly classified as $9$.]{\includegraphics[scale=0.1]{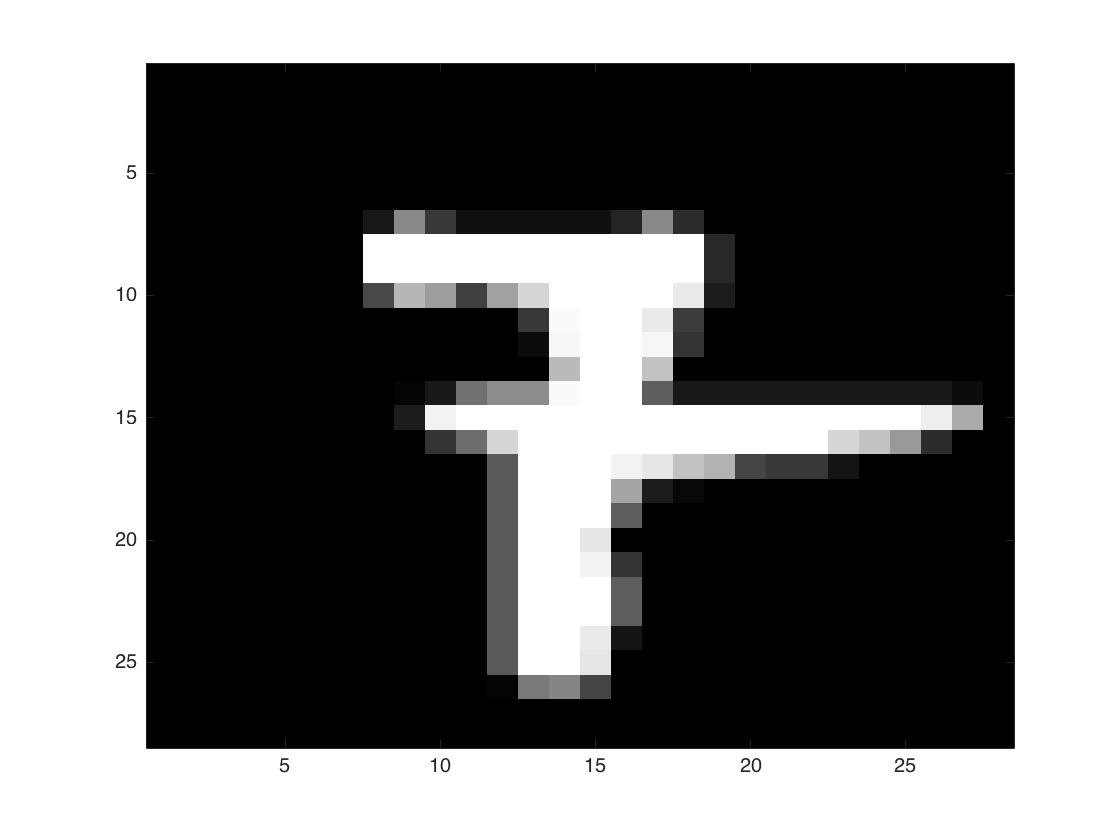}}
\subfigure[Incorrectly classified as $2$.]{\includegraphics[scale=0.1]{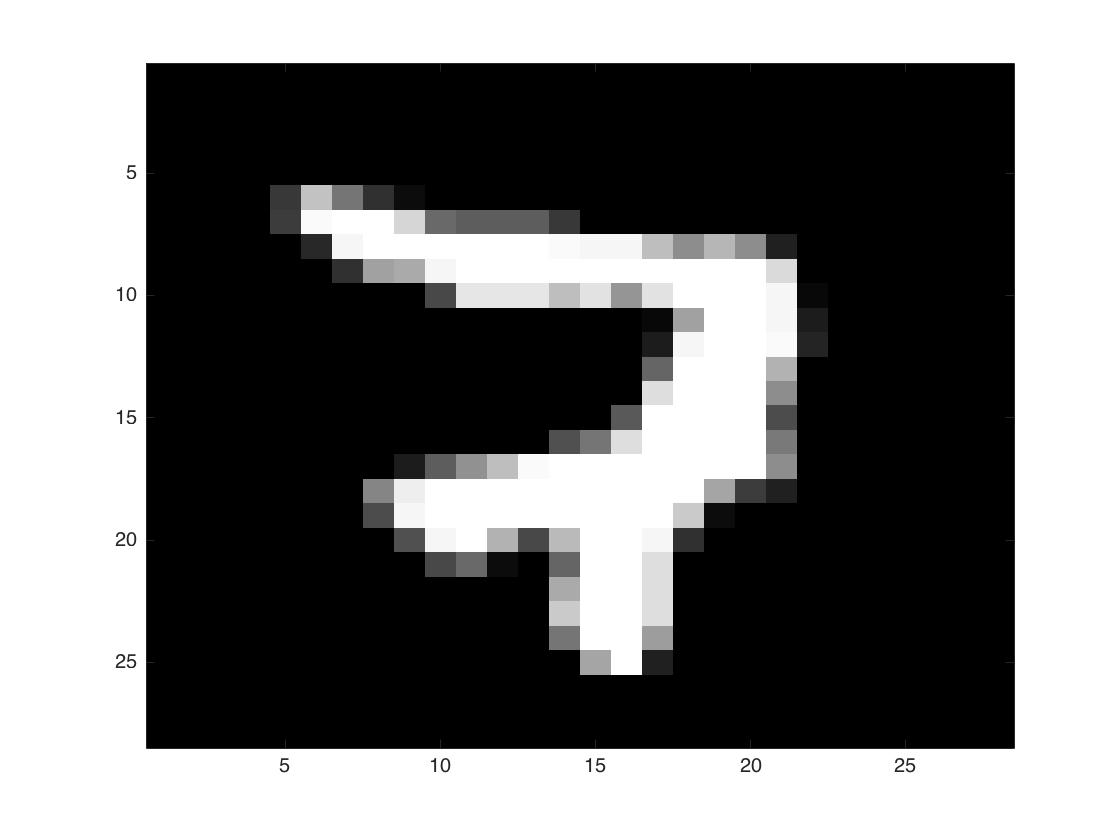}}
\caption{Examples of incorrect classification of images that should be 7.}
\label{fig:classification example}
\end{figure}

We illustrate some of the mis-classifications that occur in Figure~\ref{fig:classification example} for truncation $k = 4$. 
We notice that Figure~\ref{fig:classification example}a and \ref{fig:classification example}b do have qualitative similarities to $9$ and $2$, respectively. We can likely improve for ambiguous digits by adding additional features for each class and/or employing slightly different metrics.

%


\subsection{Numerical Results:  Identification}

Our second objective is to use our local tSVD feature vectors to determine if a pair of test images contain the same digit.  To solve this problem, we consider each comparison \eqref{defn:projection} to be a \emph{feature} for a particular image $\vec{\Pcal}_j$ instead of minimizing over the number of classes.  More specifically, we construct a $1\times 10$ vector of features for each of our 10,000 test images.

 We measure the similarity between two images by computing the cosine between the feature vectors.  Though other similarity metrics are possible, given what all the (non-negative) entries in the feature vector represent, this seemed appropriate for proof of concept. 

We compute the similarity for all $(i,j)$-pairs of test images to form a similarity score matrix $S$ of size $10000 \times 10000$ where $S$ is symmetric.
	\begin{figure}[H]
	\centering
	\caption{Similarity score matrix for truncation $k = 4$.}
	\includegraphics[scale=0.18]{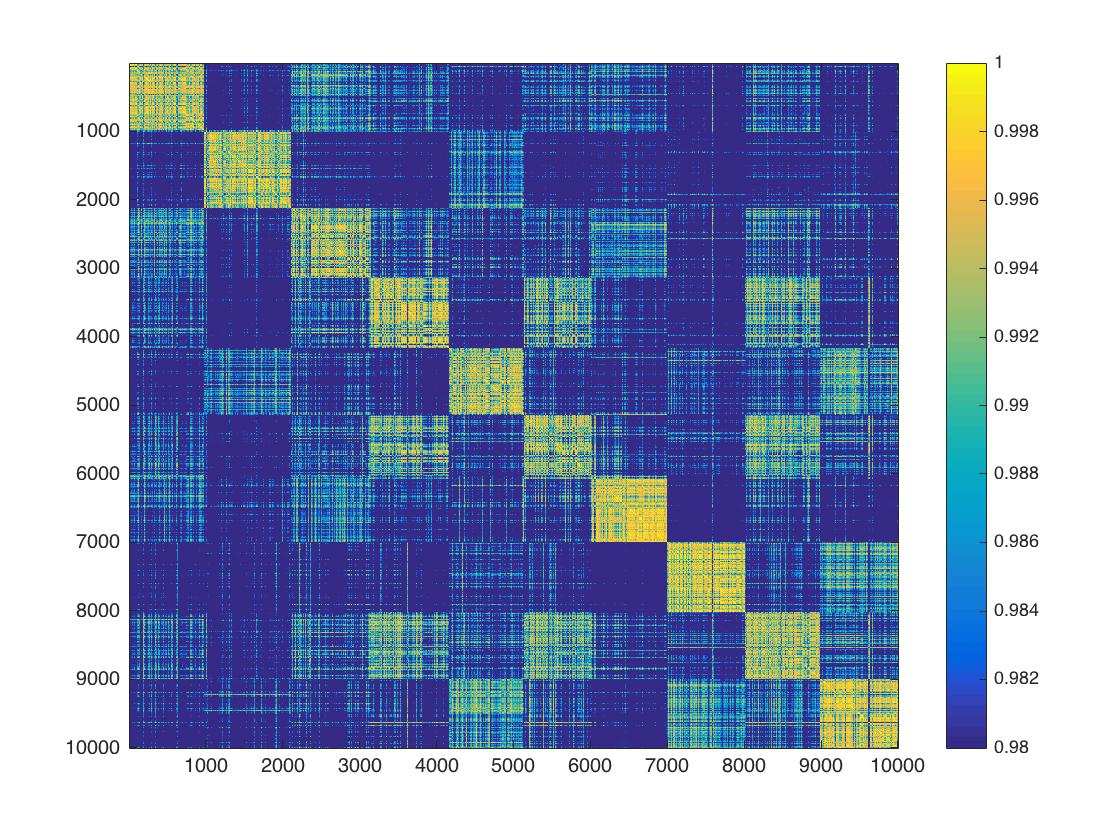}
	\label{fig:matrix k3}
	\end{figure}	
In Figure~\ref{fig:matrix k3}, we display only the similarity scores between $0.98$ and $1$ and we notice that blocks along the diagonal contain the highest similarity scores, as desired given the ordering of the test data.   
This illustrates that the cosine metric does enable us to determine if two images contain the same digit.

\begin{figure}[H]
	\centering
	\caption{ROC curve for various truncation values $k$.}
	\includegraphics[scale=0.18]{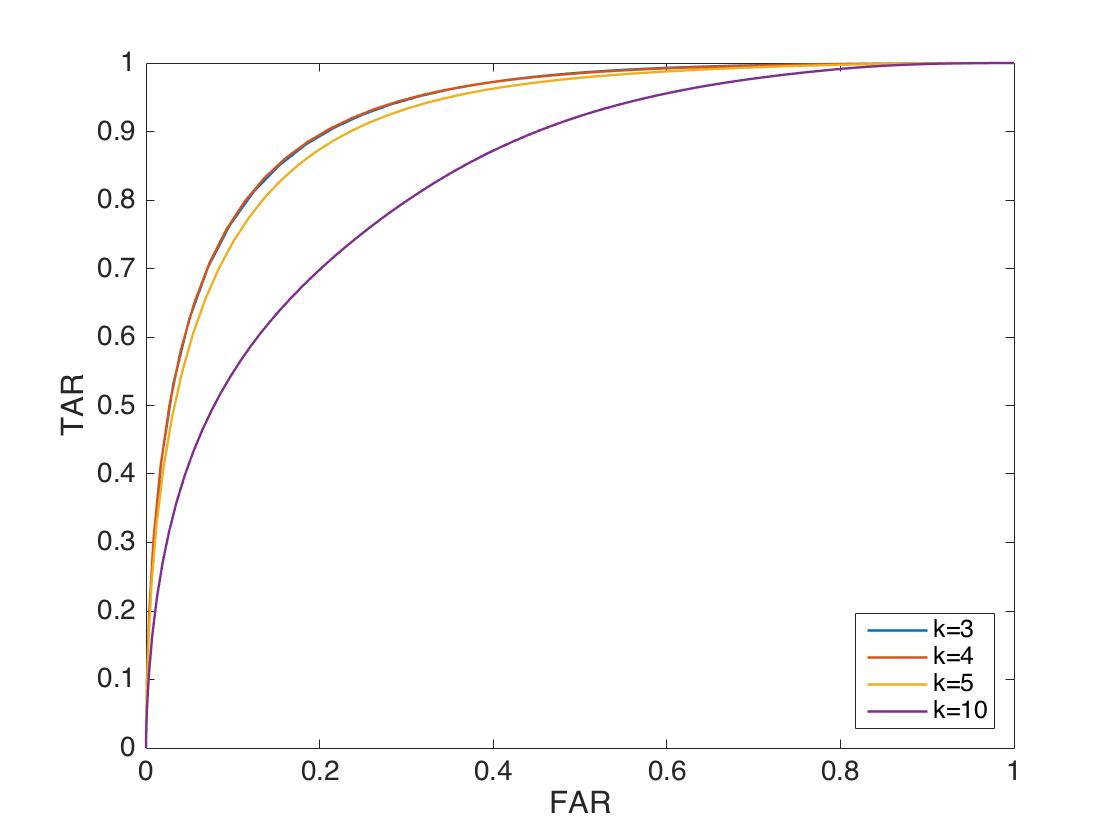}
	\label{fig:roc}
\end{figure}
	Using a receiver operating characteristic (ROC) curve in Figure~\ref{fig:roc}, we visualize the effectiveness of our local tSVD classifier. 
Notice the curve for truncation $k = 10$ is significantly lower, indicating smaller truncation values (indicative of less storage) yield better accuracy for the MNIST dataset. 

\section{Conclusions and Future Work}

We have developed a new local truncated tSVD approach to 
image classification 
based on provable optimality conditions which is elegant in its straightforward mathematical approach to the problem.   
Beyond the innate computational and storage efficiency advantages of the proposed approach, it has demonstrated effective performance 
in 
classifying MNIST data. 
The primary 
purpose of this short paper was a proof of concept of a new method. In the future, we will compare our approach to current state-of-the-art approaches in terms of storage, computation time and qualitative classification results for larger and different datasets (e.g. subjects from a dataset of facial images).  Additionally, we seek an automated strategy for determining optimal truncation value $k$ or a varied truncation scheme denoted tSVDII as in \cite{Hao2013}.  We will also explore whether the alternative tensor-tensor products from \cite{Kernfeld} and their corresponding truncated tSVDs will allow us to obtain more illustrative features, and whether new double-sided tSVD techniques \cite{JianiThesis} that are insensitive to 
tensor 
 orientation are useful here as well.


\section*{Acknowledgment}
This research is partially based upon work supported by the National Science Foundation under NSF 1319653 and by the Office of the Director of National Intelligence (ODNI), Intelligence Advanced Research Projects Activity (IARPA), via IARPA’s 2014-14071600011. The views and conclusions contained herein are those of the authors and should not be interpreted as necessarily representing the official policies or endorsements, either expressed or implied, of ODNI, IARPA, or the U.S. Government. The U.S. Government is authorized to reproduce and distribute reprints for Governmental purpose notwithstanding any copyright annotation thereon.



%

\end{document}